\documentclass[sigconf]{acmart}

\AtBeginDocument{
  \providecommand\BibTeX{{
    \normalfont B\kern-0.5em{\scshape i\kern-0.25em b}\kern-0.8em\TeX}}}

\copyrightyear{2020}
\acmYear{2020}
\setcopyright{acmcopyright}
\acmConference[MM '20]{Proceedings of the 28th ACM International Conference on Multimedia}{October 12--16, 2020}{Seattle, WA, USA}
\acmBooktitle{Proceedings of the 28th ACM International Conference on Multimedia (MM '20), October 12--16, 2020, Seattle, WA, USA}
\acmPrice{15.00}
\acmDOI{10.1145/3394171.3413802}
\acmISBN{978-1-4503-7988-5/20/10}

\acmSubmissionID{605}

\usepackage{epsfig,graphicx,hyperref,amsfonts,xcolor,multirow,amsmath,balance}
\usepackage{mycommand}

\begin{document}

\fancyhead{}

\title{Stronger, Faster and More Explainable: A Graph Convolutional Baseline for Skeleton-based Action Recognition}

\finaltrue

\iffinal

\author{Yi-Fan Song}
\email{yifan.song@cripac.ia.ac.cn}
\orcid{0000-0002-5882-6126}
\affiliation{
  \institution{$^1$University of Chinese Academy of Sciences}
  \institution{$^2$Institute of Automation, Chinese Academy of Sciences}
}

\author{Zhang Zhang}
\email{zzhang@nlpr.ia.ac.cn}
\affiliation{
  \institution{$^1$University of Chinese Academy of Sciences}
  \institution{$^2$Institute of Automation, Chinese Academy of Sciences}
}

\author{Caifeng Shan}
\email{caifeng.shan@gmail.com}
\affiliation{
  \institution{$^1$College of Electrical Engineering and Automation, Shandong University of Science and Technology}
  \institution{$^2$Artificial Intelligence Research, Chinese Academy of Sciences}
}

\author{Liang Wang}
\email{wangliang@nlpr.ia.ac.cn}
\affiliation{
  \institution{$^1$University of Chinese Academy of Sciences}
  \institution{$^2$Institute of Automation, Chinese Academy of Sciences}
  \institution{$^3$School of Computer Science and Technology, Anhui University}
}

\renewcommand{\shortauthors}{Song \etal}

\else

\author{Anonymous ACMMM submission}
\affiliation{\institution{Paper ID 605}}

\renewcommand{\shortauthors}{Anonymous}

\fi

\begin{abstract}
  One essential problem in skeleton-based action recognition is how to extract discriminative features over all skeleton joints. However, the complexity of the State-Of-The-Art (SOTA) models of this task tends to be exceedingly sophisticated and over-parameterized, where the low efficiency in model training and inference has obstructed the development in the field, especially for large-scale action datasets. In this work, we propose an efficient but strong baseline based on Graph Convolutional Network (GCN), where three main improvements are aggregated, \ie, early fused Multiple Input Branches (MIB), Residual GCN (ResGCN) with bottleneck structure and Part-wise Attention (PartAtt) block. Firstly, an MIB is designed to enrich informative skeleton features and remain compact representations at an early fusion stage. Then, inspired by the success of the ResNet architecture in Convolutional Neural Network (CNN), a ResGCN module is introduced in GCN to alleviate computational costs and reduce learning difficulties in model training while maintain the model accuracy. Finally, a PartAtt block is proposed to discover the most essential body parts over a whole action sequence and obtain more explainable representations for different skeleton action sequences. Extensive experiments on two large-scale datasets, \ie, NTU RGB+D 60 and 120, validate that the proposed baseline slightly outperforms other SOTA models and meanwhile requires much fewer parameters during training and inference procedures, \eg, at most 34 times less than DGNN, which is one of the best SOTA methods.
\end{abstract}



\maketitle

\section{Introduction}
\label{sec:introduction}

In the past decade, human action recognition becomes increasingly crucial and achieves promising progress in various applications, such as video surveillance, human-computer interaction, video retrieval and so on \cite{poppe2010survey, aggarwal2011human, weinland2011survey}. One essential problem in human action recognition is how to extract discriminative and rich features to fully describe the spatial configurations and temporal dynamics in human actions.\footnote{The codes and pretrained models of the preposed ResGCN are available at \href{http://github.com/yfsong0709/ResGCNv1}{here}.}

Currently, skeleton-based representations have been very popular for human action recognition, as human skeletons provide a compact data form to depict dynamic changes in human body movements \cite{johansson1973visual}. Skeleton data is a time series of 3D coordinates of multiple skeleton joints, which can be either estimated from 2D images by pose estimation methods \cite{cao2017realtime} or directly collected by multimodal sensors such as Kinect \cite{zhang2012microsoft}. Moreover, skeleton-based representations are more robust to the variations of illuminations, camera viewpoints and other background changes. These merits inspire researchers to develop various methods to explore informative features from skeleton motion sequences for action recognition.

The current development of skeleton-based action recognition can be divided mainly into two phrases. In early years, conventional methods adopt Recurrent Neural Network (RNN)-based or CNN-based models to analyze skeleton sequences. For example, Du \etal \cite{du2015hierarchical} employ a hierarchical bidirectional RNN to capture rich dependencies between different body parts. And Li \etal \cite{li2017skeleton} design a simple yet effective CNN architecture for action classification from trimmed skeleton sequences. In recent years, due to the greatly expressive power for depicting structural data, graph-based models \cite{kipf2016semi, li2018adaptive} have been proposed for modeling dynamic skeleton sequences. Yan \etal \cite{yan2018spatial} firstly propose the Spatial Temporal Graph Convolutional Networks (ST-GCN) for skeleton-based action recognition, after that increasing studies \cite{zhang2019graph, shi2019two, song2019richly} are reported based on GCN models.

Nevertheless, for learning discriminative and rich features from skeleton sequences, the current SOTA models are often exceedingly sophisticated and over-parameterized, where the network often contains a multi-stream architecture with a large number of model parameters, which leads to a hard training procedure and high computational costs (low inference speed). For example, the 2s-AGCN in \cite{shi2019two} contains about 6.94 million parameters, and takes nearly 4 GPU-days for model training on the NTU RGB+D 60 dataset \cite{shahroudy2016ntu}. And the DGNN \cite{shi2019skeleton} contains more than 26 million parameters which is very hard for parameter tuning on large-scale datasets. Thus, the high model complexity has seriously limited the development of skeleton-based action recognition, while there are few literatures on this issue. Moreover, the explainability issue of conventional models still lacks of considerations in current studies. Although some studies \cite{song2017end, si2019attention} have utilized attention models to discover informative skeleton joints and explain the differences between action categories, they commonly exploit the relationships among individual joints and frames, which often suffers from the noisy skeleton joints in sensor inputs or inaccurate estimations.

To tackle the above problems, we make three main improvements in this paper to build a new efficient baseline for skeleton-based action recognition. Firstly, an early fused Multiple Input Branches (MIB) architecture is proposed to capture rich spatial configurations and temporal dynamics from skeleton data, where the three branches include joint positions (relative and absolute), bone features (lengths and angles) and motion velocities (one or two temporal steps) respectively, which are subsequently fused in the early stage of the whole model for reducing the model complexity. Secondly, inspired by the success of the ResNet \cite{he2016deep} in CNN-based image classification, we introduce a Residual GCN (ResGCN) module, where the residual links make the model optimization earlier than the original unreferenced feature projection and the bottleneck structure can significantly alleviate the amount of parameter tuning costs. Finally, instead of existing joint-wise attentions in previous models, a Part-wise Attention (PartAtt) module is proposed to discover the most essential body parts over a whole action sequence, and thereby enhance the explainability and stability of the learned representations for different action sequences.

\begin{figure}[t]
  \centerline{\includegraphics[width=8cm]{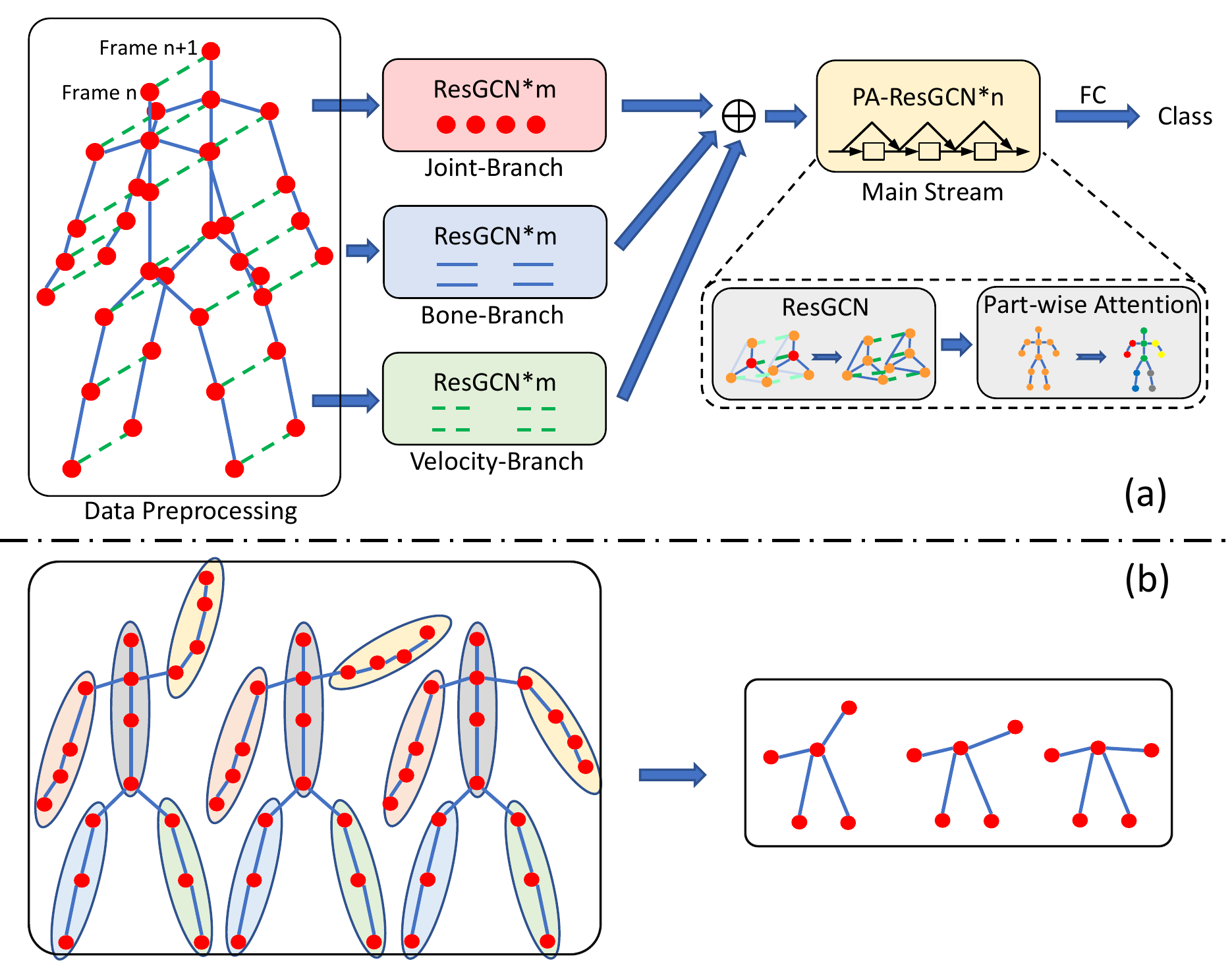}}
  \caption{(a) is the overall pipeline of our approach and (b) is an illustration of five manually designed body parts. \bv}\label{fig:pipeline}
  \Description{pipeline}
  \vspace{-0.4cm}
\end{figure}

The whole pipeline of the newly proposed baseline is shown in the top of Fig. \ref{fig:pipeline}, where the three input sequences (Joint, Velocity and Bone) are initially extracted from the original skeleton sequence. Next, each input sequence is sent to an input branch consisting of some ResGCN modules. Then, the three branches will be concatenated and sent to several PA-ResGCN modules, where each PA-ResGCN module contains a sequential execution of a ResGCN module, followed by a PartAtt block. Finally, the GCN features of all joints are concatenated and feed into a fully-connected (FC) layer for action classification. In this paper, two types of baselines are provided, \ie, a baseline with high performance (PA-ResGCN) and a baseline with high efficiency (ResGCN, without PartAtt blocks). Compared with the most popular GCN baseline, \ie, ST-GCN \cite{yan2018spatial}, the PA-ResGCN achieves over 10\% and 20\% relative performance increases with the similar model size on the two datasets, NTU RGB+D 60 \cite{shahroudy2016ntu} and 120 \cite{liu2019ntu}. Besides, the PA-ResGCN obtains the SOTA performance on the NTU 120 dataset, while it also achieves competitive performance to other SOTA methods on the NTU 60 dataset. Furthermore, when considering the model size and computational cost, the ResGCN with bottleneck structure obtains a slightly lower accuracy than other SOTA models, while it only contains 0.77 million parameters, nearly 34 times less than DGNN \cite{shi2019skeleton}, which is one of the best SOTA methods. The main contributions of the proposed baseline can be summarized as follows:
\begin{itemize}
  \item	An early fused multi-branch architecture is designed to take inputs from three individual spatio-temporal feature sequences (Joint, Velocity and Bone) obtained from raw skeleton data, which enables the baseline model to extract sufficient structural features.
  \item To further enhance the efficiency of our model, a residual bottleneck structure is introduced in GCN, where the residual links reduce the difficulties in model training and the bottleneck structure reduces the computational costs in parameter tuning and model inference.
  \item A part-wise attention block is proposed to compute attention weights for different human body parts to further improve the discriminative capability of the features, which meanwhile provides an explanation for the classification results through visualizing the class activation maps.
  \item Extensive experiments are conducted on two large-scale skeleton action datasets, \ie, NTU RGB+D 60 and 120, where the PA-ResGCN can achieve the SOTA performance, and the ResGCN with bottleneck structure obtains competitive performance with much fewer parameters.
\end{itemize}

\section{Related work}
\label{sec:related}

\paragraph{Skeleton-based Action Recognition.} Due to its compactness to the RGB-based representations, action recognition based on skeleton data has received increasing attentions. In an earlier work \cite{li2018co}, a convolutional co-occurrence feature learning framework is proposed, where a hierarchical methodology is employed to gradually aggregate different levels of contextual information. The study in \cite{zhang2019view} designs a view adaptive model to automatically regulate observation viewpoints during the occurrence of an action, so as to obtain the view invariant representations of human actions.

Inspired by the booming graph-based methods, Yan \etal \cite{yan2018spatial} firstly introduce GCN into the skeleton-based action recognition task, and propose the ST-GCN to model the spatial configurations and temporal dynamics of skeletons synchronously. Following this work, Song \etal \cite{song2019richly} aims to solve the occlusion problem in this task, and propose a multi-stream GCN to extract rich features from more activated skeleton joints. Shi \etal \cite{shi2019two} utilize a Non-local method into a two-stream GCN model, which significantly improves the model's accuracy. However, these well-performance models are usually based on multi-stream structures, which need to tune a larger amount of parameters with higher computational costs. Therefore, how to reduce the complexity of the GCN model is still a challenging problem for 3D skeleton based action recognition.

\paragraph{Efficient Models.} Some existing studies have been considering the model complexity problem. The study \cite{yang2019make} constructs a lightweight network with CNN-based blocks, which is not as accurate as GCN models. The work \cite{zhang2019semantics} adopts a complex data preprocessing strategy, whose inputs include positions, velocities, frame indexes and joint types. This data preprocessing module enables the model to recognize actions with a shallow model, thereby achieves a very fast inference speed with 188 sequences/(second*GPU), yet its performance is obviously lower than other SOTA models.

\paragraph{Attention Models.} Attention mechanisms have become an integral part of compelling sequence modeling in various tasks, such as action recognition. Baradel \etal \cite{baradel2017human} introduce the attention mechanism into an RGB-based action recognition model, which uses human pose to calculate spatial and temporal attentions. The study in \cite{song2017end} firstly introduces attention modules into skeleton-based action recognition, where a spatial-temporal attention Long Short-Term Memory (LSTM) is built to allocate different levels of attention to the discriminative joints within each frame. Si \etal \cite{si2019attention} also incorporate attention modules within LSTM units. Both the two models apply attention modules for each frame individually, which may attend to some unstable noisy features. Besides, the traditional attention module is usually implemented by a multi-layer perception, which does not consider the intensive local dependency for temporal attention and the part dependency for spatial attention.

\paragraph{Part-based Models.} Human skeleton is a natural graph with five main body parts, as shown in Fig. \ref{fig:pipeline}. Thus, part-based methods are often designed by researchers to extract the features of body parts individually. Du \etal \cite{du2015hierarchical} propose a bidirectional RNN to hierarchically concatenate the features of body parts. Thakkar \etal \cite{thakkar2018part} utilize GCN to model different body parts, then aggregate them together to recognize actions. Recently, Huang \etal \cite{huang2020part} propose a part-based skeleton model, which is capable to synchronously explore discriminative features from joints and body parts. All these part-based models aim to extract features from body parts individually, while our work focuses on discovering the most informative parts with attention mechanisms.

\begin{figure}
  \centerline{\includegraphics[width=8.5cm]{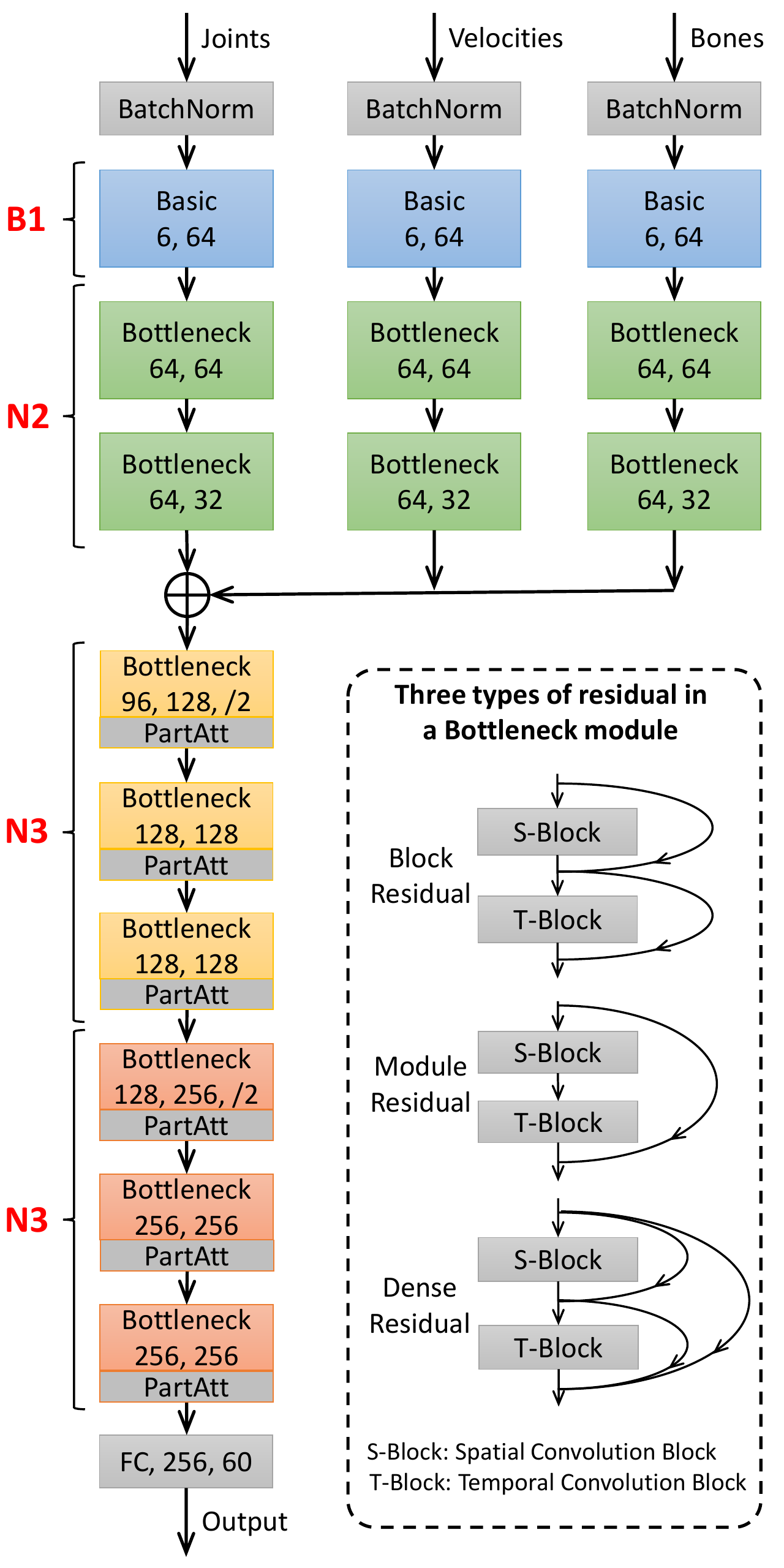}}
  \caption{An example of the ResGCN model with bottleneck structure. The structural parameters of this example are {\bf [B1,N2,N3,N3]}, which correspond to the type and the number of modules in different model parts. Concretely, B1 denotes one ResGCN module with basic (B) blocks and N2/N3 denotes two/three ResGCN modules with bottleneck (N) blocks. Each module in the network is composed of a spatial block, a temporal block and a residual link. In addition, this figure illustrates three types of residual links, \ie, {\bf Block} residual, {\bf Module} residual and {\bf Dense} residual, shown at the bottom-right corner. \bv}\label{fig:model}
  \Description{model}
\end{figure}

\section{Methods}
\label{sec:methods}

In this section, we will illustrate the proposed ResGCN/PA-ResGCN. Firstly, the GCN operation will be briefly introduced. Next, we will discuss the details of ResGCN, which can be constructed by stacking some basic or bottleneck blocks. Then, the multiple input branches (MIB) with data preprocessing will be presented. Finally, the new PartAtt block is proposed to enhance the model performance and explainability. An example of the baseline with bottleneck structure is displayed in Fig. \ref{fig:model}.

\subsection{Graph Convolutional Network}
\label{ssec:gcn}

According to \cite{yan2018spatial}, the spatial GCN operation for each frame $t$ in a skeleton sequence is formulated as
\begin{equation}\label{eq:gcn}
f_{out}=\sum_{d=0}^{D} W_d f_{in} (\Lambda_d^{-\frac{1}{2}}A_d\Lambda_d^{-\frac{1}{2}} \otimes M_d),
\end{equation}
where $D$ is a predefined maximum graph distance, $f_{in}$ and $f_{out}$ denote the input and output feature maps, $\otimes$ means element-wise multiplication, $A_d$ represents the $d$-th order adjacency matrix that marks the pairs of joints with a graph distance $d$, and $\Lambda_d$ is used to normalize $A_d$. $W_d$ and $M_d$ are both learnable parameters, which are utilized to implement the convolution operation and tune the importance of each edge, respectively.

For temporal feature extraction, an $L\times1$ convolutional layer is designed to aggregate the contextual features embedded in adjacent frames. In this operation, $L$ is a predefined hyper-parameter, defining the length of temporal windows. Both spatial and temporal convolutional layers are followed by a BatchNorm layer and a ReLU layer, and totally construct a basic block.

\subsection{ResGCN}
\label{ssec:details}

\paragraph{Bottleneck.} He \etal \cite{he2016deep} suggest a subtle block structure named bottleneck, which inserts two $1\times1$ convolutional layers before and after the common convolution layer, respectively, in order to reduce the number of feature channels with a reduction rate $r$ in convolution calculation.

In this paper, we replace spatial and temporal basic blocks with the bottleneck structure, and obtains a significantly faster implementation of model training and inference. Suppose that the input and output channels are both 256, and the channel reduction rate $r$ is 4, the temporal window size $L$ is 9. Then, the basic block contains $256\times256\times9=589824$ parameters, while the bottleneck block only contains $256\times64+64\times64\times9+64\times256=69632$ parameters, nearly 8.5 times less than the basic block. In Fig. \ref{fig:model}, each module in ResGCN contains a sequential execution of one spatial block and one temporal block respectively.

\paragraph{Residual Links.} Based on the spatial and temporal blocks mentioned above, it is easy to construct a ResGCN module after adding residual links over the blocks. There are three types of residual links, \ie, block, module and dense, displayed in the bottom-right of Fig. \ref{fig:model}. As we can see, the block residual link connects the features before and after each block, while the module link jumps the whole module. It seems that the dense link possesses both advantages of the other two links, but more links may harm the compactness of the model and need more memory costs. Therefore, it is necessary to select an appropriate type of residual links.

\subsection{Multiple Input Branches}
\label{ssec:branches}

\paragraph{Model Architecture.} Fig. \ref{fig:model} gives an example of the MIB architecture, which can be summarized by a set of hyper-parameters [B1,N2,N3,N3]. The first parameter denotes that we use one ResGCN module with basic (B) blocks to process the initial input data. And the other three parameters represent the ResGCN modules with bottleneck (N) blocks, while the differences locate at the number of input and output channels. Every ResGCN module in the third and the fourth parts is followed by a PartAtt block. In addition, at the beginning module of the third and the fourth parts, a temporal stride of 2 is used to further reduce the complexity, which is also found useful for avoiding over-fitting in our experiments.

Furthermore, it should be noticed that current high accuracy models usually apply a multi-stream architecture with various input data. For example, Shi \etal \cite{shi2019two} take the joints data and bones data as input for feeding to the same model separately, and eventually choose the fusion results of two streams as the final decision. This is an effective way to augment the input data and enhance the model performance. However, a multi-stream network often means high computation costs and difficulties of parameter turning on large-scale datasets. Thus, we fuse the three input branches at the early stage of our model, and apply one main stream to extract discriminative features after the concatenation of three branches of features. This architecture not only retains the rich input features, but also significantly suppresses the model complexity, and makes the training procedure easier to converge.

\begin{figure*}
  \centerline{\includegraphics[width=18cm]{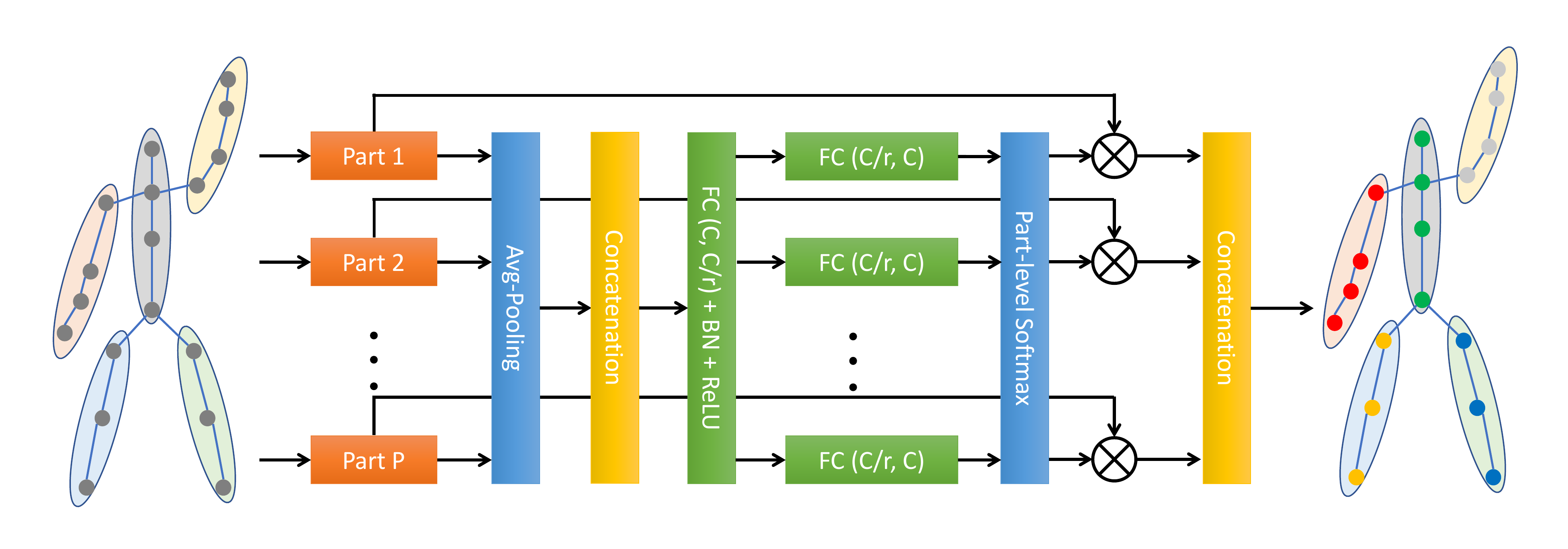}}
  \vspace{-0.4cm}
  \caption{The overview of the proposed PartAtt block, where $C$ denotes the number of input channels, $r=4$ is utilized to reduce the computational cost, $P=5$ represents five individual body parts, $\otimes$ means the element-wise multiplication and Part-level Softmax means to calculate Softmax at part level. \bv}\label{fig:part}
  \Description{part}
\end{figure*}

\paragraph{Data Preprocessing.} Data preprocessing is essential for skeleton-based action recognition, according to the previous studies \cite{song2019richly, si2018skeleton, shi2019two}. In this work, the input features after various preprocessing methods are mainly divided into three classes: {\bf 1)} joint positions, {\bf 2)} motion velocities and {\bf 3)} bone features.

Suppose that the original 3D coordinate set of an action sequence is $\mathcal{X}=\{x\in \mathbb{R}^{C\times T\times V}\}$, where $C$, $T$, $V$ denote the coordinate, frame and joint, respectively. Then the relative position set $\mathcal{R}=\{r_{i}|i=1,2,\cdots,V\}$, where $r_{i}=x[:,:,i]-x[:,:,c]$, $x[:,:,c]$ represents the center joint of a skeleton (center spine). These two sets are concatenated into a single sequence, and sent to the first branch as the input of joint positions. Moreover, it is easy to obtain the two sets of motion velocities $\mathcal{F}=\{f_t|t=1,2,\cdots,T\}$ and $\mathcal{S}=\{s_t|t=1,2,\cdots,T\}$ with the following definitions: $f_t=x[:,t+2,:]-x[:,t,:]$ and $s_t=x[:,t+1,:]-x[:,t,:]$. And the input of motion velocities is acquired by concatenating $\mathcal{F}$ and $\mathcal{S}$ for each joint to obtain a 6-d feature vector at each time. Finally, the input of bone features consists of the bone lengths $\mathcal{L}=\{l_i|i=1,2,\cdots,V\}$ and the bone angles $\mathcal{A}=\{a_i|i=1,2,\cdots,V\}$. To obtain these two sets, the displacement of each bone is calculated by $l_i=x[:,:,i]-x[:,:,i_{adj}]$, where $i_{adj}$ means the adjacent joint of the $i$-th joint. Next, the angle of each bone is calculated by
\begin{equation}
a_{i,w}=arccos(\frac{l_{i,w}}{\sqrt{l_{i,x}^2+l_{i,y}^2+l_{i,z}^2}}),
\end{equation}
where $w\in\{x,y,z\}$ denotes the 3D coordinates.

\subsection{Part-wise Attention}
\label{ssec:pa}

Previous part-based models usually aim to extract feature from body parts individually, while we focus on discovering the importance of different body parts over the whole action sequences. Inspired by Split Attention (SplitAtt) in the ResNeSt model \cite{resnest2020zhang}, the PartAtt block is designed as Fig. \ref{fig:part}. Firstly, five individual body parts are obtained from the input features by manually selecting corresponding joints (seen as Fig. \ref{fig:pipeline}). Then, the features of all parts are concatenated and average pooled in temporal dimension, and then passed through a fully connected layer with a BatchNorm layer and a ReLU function. Subsequently, five fully connected layers are adopted to calculate the attention matrices and a Softmax function is utilized to determine the most essential body parts. Finally, the features of five parts are concatenated as an integral skeleton representation with different attention weights. This PartAtt block can be formulated as
\begin{equation}
f_{p}=f_{in}(p)\otimes\delta(\theta(pool(f_{in})W)W_p)
\end{equation}
\begin{equation}
f_{out}=Concat(\{f_p|p=1,2,\cdots,P\})
\end{equation}
where $f_{in}$ and $f_{out}$ denote input and output feature maps, $\otimes$ means element-wise multiplication, $pool(\cdot)$ denotes temporal avg-pool and part-pool operations, $\delta(\cdot)$ and $\theta(\cdot)$ represent part-level Softmax and ReLU activation functions. And $W\in\mathbb{R}^{C\times\frac{C}{r}}$, $W_p\in\mathbb{R}^{\frac{C}{r}\times C}$ are both learnable parameters, where $W$ is shared by all parts for dimension reduction and $W_p$ is specific to each part for calculating the final attention weights.

The main difference between PartAtt and SplitAtt is that the cardinal groups of SplitAtt are obtained by separating feature channels, while the cardinal groups in PartAtt correspond to various body parts from the spatial view. Compared to other attention models \cite{song2017end, si2019attention}, there are two obvious differences between our PartAtt and their methods. On one hand, we employ this block to work on body parts, while their attention blocks concentrate on joints. On the other hand, traditional spatial attention blocks work for each frame individually, while our spatial attention block is based on the global contextual feature maps obtained by average pooling over the whole temporal sequence.

\section{Experimental Results}
\label{sec:experiments}

In this section, we evaluate the performance of the proposed PA-ResGCN and ResGCN (without PartAtt blocks) on two large-scale datasets NTU RGB+D 60 \cite{shahroudy2016ntu} and NTU RGB+D 120 \cite{liu2019ntu}. Ablation studies are also performed to validate the contributions of each component in our model. For simplicity, all experiments in ablation studies choose ResGCN with [B1,N2,N3,N3] structure as the base model (seen as Fig. \ref{fig:model}), which can be denoted as ResGCN-N51, where N51 means there are 51 convolutional or FC layers within the model. Similarly, the model with the same architecture and basic blocks can be denoted as ResGCN-B19. Finally, result analyses are reported to prove the effectiveness of the proposed PartAtt block.

\begin{table*}
  \begin{center}
  \begin{tabular}{cc|cc|cc|cc}
  \hline
  Model & Conf. & Speed & Param. & X-sub & X-view & X-sub120 & X-set120 \\
  \hline
  \hline
  HBRNN \cite{du2015hierarchical} & CVPR15 & -- & -- & 59.1 & 64.0 & -- & -- \\
  ST-LSTM \cite{liu2016spatio} & ECCV16 & -- & -- & 69.2 & 77.7 & 55.0 & 57.9 \\
  TSRJI \cite{caetano2019skeleton} & SIBGRAPI19 & -- & -- & 73.3 & 80.3 & 67.9 & 62.8 \\
  TSA \cite{caetano2019skelemotion} & AVSS19 & -- & -- & 76.5 & 84.7 & 67.7 & 66.9 \\
  VA-fusion \cite{zhang2019view} & TPAMI19 & -- & 24.60 & 89.4 & 95.0 & -- & -- \\
  \hline
  \hline
  ST-GCN \cite{yan2018spatial} & AAAI18 & 42.9$^\star$ & 3.10$^\star$ & 81.5 & 88.3 & 70.7$^\dagger$ & 73.2$^\dagger$ \\
  SR-TSL \cite{si2018skeleton} & ECCV18 & 14.0$^\star$ & 19.07$^\star$ & 84.8 & 92.4 & -- & -- \\
  PB-GCN \cite{thakkar2018part} & BMVC18 & -- & -- & 87.5 & 93.2 & -- & -- \\
  RA-GCN \cite{song2019richly} & ICIP19 & 18.7$^\dagger$ & 6.21$^\star$ & 85.9 & 93.5 & 74.6$^\dagger$ & 75.3$^\dagger$ \\
  GR-GCN \cite{gao2019optimized} & ACMMM19 & -- & -- & 87.5 & 94.3 & -- & -- \\
  AS-GCN \cite{li2019actional} & CVPR19 & -- & 6.99$^\star$ & 86.8 & 94.2 & 77.9$^\dagger$ & 78.5$^\dagger$ \\
  2s-AGCN \cite{shi2019two} & CVPR19 & 22.3$^\dagger$ & 6.94$^\star$ & 88.5 & 95.1 & 82.5$^\dagger$ & 84.2$^\dagger$ \\
  AGC-LSTM \cite{si2019attention} & CVPR19 & -- & 22.89$^\star$ & 89.2 & 95.0 & -- & -- \\
  DGNN \cite{shi2019skeleton} & CVPR19 & -- & 26.24$^\star$ & 89.9 & {\bf 96.1} & -- & -- \\
  AS-GCN+DH-TCN \cite{papadopoulos2019vertex} & arXiv19 & -- & -- & 85.3 & 92.8 & 78.3 & 79.8 \\
  SGN \cite{zhang2019semantics} & arXiv19 & {\bf 188.0} & 1.8 & 86.6 & 93.4 & -- & -- \\
  PL-GCN \cite{huang2020part} & AAAI20 & -- & 20.70$^\star$ & 89.2 & 95.0 & -- & -- \\
  NAS-GCN \cite{peng2020learning} & AAAI20 & -- & 6.57$^\star$ & 89.4 & 95.7 & -- & -- \\
  \hline
  \hline
  ResGCN-N51 (Bottleneck) & ours & {\bf 67.4} & {\bf 0.77} & 89.1 & 93.5 & 84.0 & 84.2 \\
  PA-ResGCN-N51 & ours & 54.8 & 1.14 & 90.3 & 95.6 & 86.6 & 87.1 \\
  \hline
  ResGCN-B19 (Basic) & ours & 44.0 & 3.26 & 90.0 & 94.8 & 85.2 & 85.7 \\
  PA-ResGCN-B19 & ours & 38.3 & 3.64 & {\bf 90.9} & {\bf 96.0} & {\bf 87.3} & {\bf 88.3} \\
  \hline
  \multicolumn{8}{l}{$^\star$: These results are provided by the authors or calculated according to their released codes.}\\
  \multicolumn{8}{l}{$^\dagger$: These results are implemented by ourselves, based on their released codes on the Github website.}\\
  \end{tabular}
  \end{center}
  \caption{Comparison with the SOTA methods on NTU RGB+D 60 \& 120 datasets in accuracy (\%), inference speed (sequences/(second*GPU)) and parameter number (million). The top part consists of several models without the GCN technique, while the middle part contains some graph-based models.}\label{tab:ntu}
  \vspace{-0.6cm}
\end{table*}

\subsection{Datasets}
\label{ssec:datasets}

\paragraph{NTU RGB+D 60 Dataset.} This large-scale indoor captured dataset is provided in \cite{shahroudy2016ntu}, which contains 56680 human action videos collected by three Kinect v2 cameras. These actions consist of 60 classes, where the last 10 classes are all interactions between two subjects. For simplicity, the input frame number is set to 300, and the sequences with less than 300 frames are padded by 0 at the end. Each frame contains no more than 2 skeletons, and each skeleton is composed of 25 joints. The authors of this dataset recommend two benchmarks: {\bf 1) cross-subject (X-sub)} contains 40320 training videos and 16560 evaluation videos divided by splitting the 40 subjects into two groups. {\bf 2) cross-view (X-view)} recognizes the videos collected by cameras 2 and 3 as training samples (37920 videos), while the videos collected by camera 1 are treated as evaluation samples (18960 videos). Note that there are 302 wrong samples selected by \cite{liu2019ntu} that need to be ignored.

\paragraph{NTU RGB+D 120 Dataset.} This is the currently largest indoor action recognition dataset \cite{liu2019ntu}, which is an extended version of the NTU RGB+D 60. It totally contains 114480 videos performed by 106 subjects from 155 viewpoints. These videos consist of 120 classes, extended from the 60 classes of the previous dataset. Similarly, two benchmarks are suggested: {\bf 1) cross-subject (X-sub120)} is divided subjects into two groups, to construct training and evaluation sets (63026 and 50922 videos respectively). {\bf 2) cross-setup (X-set120)} contains 54471 videos for training and 59477 videos for evaluation, which are separated based on the distance and height of their collectors. According to \cite{liu2019ntu}, 532 bad samples of this dataset should be ignored in all experiments.

\subsection{Implementation Details}
\label{ssec:Implementation}

In our experiments, the maximum graph distance $D$ and the temporal window size $L$ mentioned in Section \ref{ssec:gcn} are set to 2 and 9, respectively. The maximum number of training epochs is set to 70. The initial learning rate is set to 0.1 and decays by 10 at the 20-th and 50-th epochs. Moreover, a warmup strategy \cite{he2016deep} is utilized at the first 10 epochs to make the training procedure more stable. The stochastic gradient descent (SGD) with the Nesterov momentum of 0.9 and the weight decay of 0.0001 is employed to tune the parameters. Other structural parameters are defined as Fig. \ref{fig:model}. In addition, the dropout layer in original ST-GCN model \cite{yan2018spatial} is removed. All our experiments are performed on two GTX TITAN X GPUs.

\subsection{Comparisons with SOTA Methods}
\label{ssec:comparisons}

\subsubsection{NTU RGB+D 60 Dataset}
\label{sssec:ntu60}

\paragraph{vs. SOTA Models.} From Tab. \ref{tab:ntu}, the PA-ResGCN-B19 obtains an excellent performance, 90.9\% for X-sub benchmark and 96.0\% for X-view benchmark. When removing the PartAtt blocks and replacing basic blocks with bottleneck blocks, the ResGCN-N51 is built, and its recognition accuracies are 89.1\% and 93.5\% for the two benchmarks, respectively, but with only a quarter amount of model parameters compared to PA-ResGCN-B19. Here, three typical methods should be noticed. {\bf 1)} The first one is ST-GCN \cite{yan2018spatial}, which is the currently most popular backbone model for skeleton-based action recognition. Compared with ST-GCN, our ResGCN-N51 outperforms by 7.6\% on X-sub benchmark and 5.2\% on X-view benchmark. {\bf 2)} 2s-AGCN \cite{shi2019two} is another popular baseline in skeleton-based action recognition. The proposed baseline ResGCN-N51 outperforms 2s-AGCN in both accuracy and efficiency. {\bf 3)} The third one is DGNN \cite{shi2019skeleton}, which is the current SOTA method with GCN technique. The ResGCN-N51 is slightly lower than DGNN in accuracy, while ResGCN-N51 only requires 0.77 million parameters, about 34 times less than that of DGNN. With respect to PA-ResGCN-B19, our model achieves SOTA performance on NTU 60 dataset with only 1/8 parameters of DGNN. This gap of model complexity is caused by the multi-stream architecture in DGNN, while ResGCN only contains one main stream. These results imply that the proposed ResGCN is an efficient baseline with competitive performance to SOTA methods.

\paragraph{vs. Efficient Models.} In order to verify the efficiency of our model with bottleneck blocks, we compare ResGCN-N51 with other methods in accuracy and inference speed on X-sub benchmark. The inference speed is defined as the number of sequences successfully evaluated by the model in one second with one GPU. The inference speeds are demonstrated in Tab. \ref{tab:ntu}. From this table, it can be found that ResGCN-N51 greatly improves the inference speed compared with the ST-GCN model \cite{yan2018spatial}. For LSTM-based models such as VA-fusion \cite{zhang2019view} and SR-TSL \cite{si2018skeleton}, the inference speeds are very slow, because of the high computational cost of the LSTM technique. SGN \cite{zhang2019semantics} is a lightweight model which contains five GCN or CNN layers and obtains an extremely fast inference speed. However, it only achieves an accuracy of 86.6\% on X-sub benchmark, significantly worse than the ResGCN-N51. Therefore, the ResGCN-N51 is a considerable model that balances the performance and efficiency.

\paragraph{vs. Attention Enhanced Models.} STA-LSTM \cite{song2017end} and AGC-LSTM \cite{si2019attention} are also enhanced by attention blocks. However, there are obvious differences between PA-ResGCN and these two models, \eg, our attention block works for the whole sequence and body part, while their models use the attention block for each frame and each joint. The performance of PA-ResGCN greatly exceeds STA-LSTM over 10\% on the two benchmarks, and outperforms AGC-LSTM by 1.7\% and 1.0\% on X-sub and X-view benchmarks, respectively.

\paragraph{vs. Part-based Models.} There are three part-based models, \ie, HBRNN \cite{du2015hierarchical}, PB-GCN \cite{thakkar2018part} and PL-GCN \cite{huang2020part}, which often incorporate the part-based operation into the RNN or GCN blocks for a more informative representation, while PA-ResGCN calculates part-wise attentions to discover the key body parts. As shown in Tab. \ref{tab:ntu}, the proposed PA-ResGCN outperforms PL-GCN in terms of the two benchmarks. As to HBRNN and PB-GCN, our approach has superior performance because the model architectures of HBRNN and PB-GCN are too simple to sufficiently explore discriminative features.

\subsubsection{NTU RGB+D 120 Dataset}
\label{sssec:ntu120}

The NTU RGB+D 120 dataset is proposed by Liu \etal \cite{liu2019ntu} recently. As a newly released dataset, there is a little work performed on the new dataset. For more convincing, four popular models, \ie, ST-GCN \cite{yan2018spatial}, RA-GCN \cite{song2019richly}, AS-GCN \cite{li2019actional} and 2s-AGCN \cite{shi2019two}, are implemented by ourselves, based on their released codes. The right column of Tab. \ref{tab:ntu} presents the experimental results, from which we can find a huge gap between the proposed ResGCN/PA-ResGCN and other models. For example, PA-ResGCN-B19 outperforms 2s-AGCN by 4.8\% and 4.1\% for the two benchmarks. We consider that this phenomenon is caused by the capability of PartAtt blocks to discovering features from the most informative body parts.

\subsection{Ablation Studies}
\label{ssec:ablation}

\begin{table}
  \begin{center}
  \begin{tabular}{c|cc|c}
  \hline
  & Setting & Param. & X-sub \\
  \hline
  \hline
  & ResGCN-B19 (Basic) & 3.35 & {\bf 90.0} \\
  Block & ResGCN-N51 (Bottleneck) ($r=2$) & 1.61 & 89.0 \\
  & ResGCN-N51 (Bottleneck) ($r=4$) & 0.77 & {\bf 89.1} \\
  & ResGCN-N51 (Bottleneck) ($r=8$) & 0.49 & 87.2 \\
  \hline
  \hline
  & ResGCN-N51 w/o Residual & 0.63 & 85.3 \\
  Residual & + Block Residual & 0.77 & {\bf 89.1} \\
  Link & + Module Residual & 0.69 & 87.8 \\
  & + Dense Residual & 0.82 & 88.9 \\
  \hline
  \end{tabular}
  \end{center}
  \caption{Comparison with different types of blocks and residual links on X-sub benchmark in accuracy (\%) and parameter number (million). $r$ means the reduction rate.}\label{tab:setting}
  \vspace{-0.4cm}
\end{table}

\begin{table}
  \begin{center}
  \begin{tabular}{c|cc|c}
  \hline
  Architecture & Input data & Param. & X-sub \\
  \hline
  \hline
  Three-branch & ResGCN-N51 & 0.77 & {\bf 89.1} \\
  \hline
  & w/o Joint & 0.71 & 88.2 \\
  Two-branch & w/o Velocity & 0.71 & 88.0 \\
  & w/o Bone & 0.71 & 86.7 \\
  \hline
  & w/o Joint \& Bone & 0.67 & 86.6 \\
  One-branch & w/o Bone \& Velocity & 0.67 & 86.1 \\
  & w/o Joint \& Velocity & 0.67 & 84.5 \\
  \hline
  \end{tabular}
  \end{center}
  \caption{Comparison with different input data on X-sub benchmark in accuracy (\%) and parameter number (million).}\label{tab:input}
  \vspace{-0.4cm}
\end{table}

\paragraph{Bottleneck Block.} In Section \ref{ssec:details}, we introduce the bottleneck structure into the GCN model, for reducing the model size and computational cost. There is a hyper-parameter in the bottleneck structure, \ie, the reduction rate $r$, which determines the number of channels in middle layers. The top part of Tab. \ref{tab:setting} illustrates the influence of the bottleneck structure, from which our ResGCN with basic blocks achieves an excellent performance on X-sub benchmark. After introducing the bottleneck structure, the performance of our model is slightly decreased by about 1\%. Except the very large reduction rate ($r=8$), the ResGCN-N51 obtains competitive accuracies but only with a half or even a quarter amount of model parameters, compared to the ResGCN-B19 model. These results clearly indicate that the bottleneck structure with a proper reduction rate ($r=4$) can reduce the model complexity effectively while maintains the model accuracy.

\paragraph{Residual Links.} From Fig. \ref{fig:model}, three types of residual links are demonstrated. The bottom part of Tab. \ref{tab:setting} displays the recognition accuracies of different residual links. As shown in Tab. \ref{tab:setting}, the ResGCN-N51 with block residual link achieves the best performance, while the model with the module residual link obtains the worst accuracy. As to the dense residual link, it is not as accurate as expected, which implies that the uses of the block residual link and the module residual link simultaneously may produce somehow inconsistency in feature learning. Therefore, we use the block residual link to construct the proposed model.

\paragraph{MIB Module.} The proposed model contains three input branches, which are defined in Section \ref{ssec:branches}. Tab. \ref{tab:input} presents the ablation studies of the input data. As shown in the table, the models with only one input branch are significantly worse than the others. In contrast, the model with all input data gets the best accuracy. This implies that each input branch is necessary to the model, and our model takes a huge benefit from the MIB architecture.

\begin{table}
  \begin{center}
  \begin{tabular}{cc|cc}
  \hline
  Architecture & Param. & X-sub & X-sub120 \\
  \hline
  \hline
  ResGCN-N39 [B1,N2,N2,N2] & 0.61 & 88.7 & 83.7 \\
  ResGCN-N51 [B1,N2,N3,N3] & 0.77 & {\bf 89.1} & 84.0 \\
  ResGCN-N57 [B1,N3,N4,N3] & 0.83 & 88.6 & 84.0 \\
  ResGCN-N75 [B1,N3,N6,N4] & 1.01 & 88.7 & {\bf 84.2} \\
  \hline
  \end{tabular}
  \end{center}
  \caption{Comparison with different model structures on X-sub and X-sub120 benchmarks in accuracy (\%) and parameter number (million). ResGCN-Nx means that this model contains x conventional layers within the bottleneck blocks.}\label{tab:structure}
  \vspace{-0.4cm}
\end{table}

\begin{table}
  \begin{center}
  \begin{tabular}{cc|cc}
  \hline
  Model & Param. & X-sub & X-sub120 \\
  \hline
  \hline
  ResGCN-N51 & 0.77 & 89.1 & 84.0 \\
  \hline
  + ChannelAtt & 0.89 & 89.1 & 85.4 \\
  + FrameAtt & 0.77 & 88.6 & 84.8 \\
  + JointAtt & 0.77 & 89.1 & 85.3 \\
  + PartAtt & 1.14 & {\bf 90.3} & {\bf 86.6} \\
  \hline
  \end{tabular}
  \end{center}
  \caption{Comparison with different attentions on X-sub and X-sub120 benchmarks in accuracy (\%) and parameter number (million).}\label{tab:attention}
  \vspace{-0.4cm}
\end{table}

\paragraph{Model Architecture.} Previous studies \cite{he2016deep} tell us that a deeper model usually means a better performance, as well as a harder training procedure. In this part, we will discuss which model architecture has high performance/cost ratio. Tab. \ref{tab:structure} gives the accuracies of four example models. It is described that the ResGCN-N51 achieves the best performance on X-sub benchmark, while the ResGCN-N75 obtains the best accuracy on the larger NTU 120 dataset. The structural parameters of ResGCN-N51 are [B1,N2,N3,N3], as presented in Fig. \ref{fig:model}. This phenomenon that the deeper network does not bring the better performance here is mainly due to the limitation of variations in the NTU 60 dataset. A bigger and more difficult dataset may need a deeper ResGCN model, such as the NTU 120 dataset.

\paragraph{PartAtt Block.} To illustrate the advantages of PartAtt blocks, we design three comparative blocks, \ie ChannelAtt, FrameAtt and JointAtt, according to previous studies \cite{song2017end, woo2018cbam}. Experimental results in Tab. \ref{tab:attention} show that the proposed PartAtt achieves the best accuracies on both benchmarks. Especially on the larger benchmark X-sub120, the gaps between PartAtt and other attention blocks are more obvious. This is mainly because the PartAtt is more robust to the noisy skeleton joints in sensor input or inaccurate pose estimations.

\subsection{Discussion and Failure Cases}
\label{ssec:discuss}

\begin{figure}[t]
  \centerline{\includegraphics[width=8.5cm]{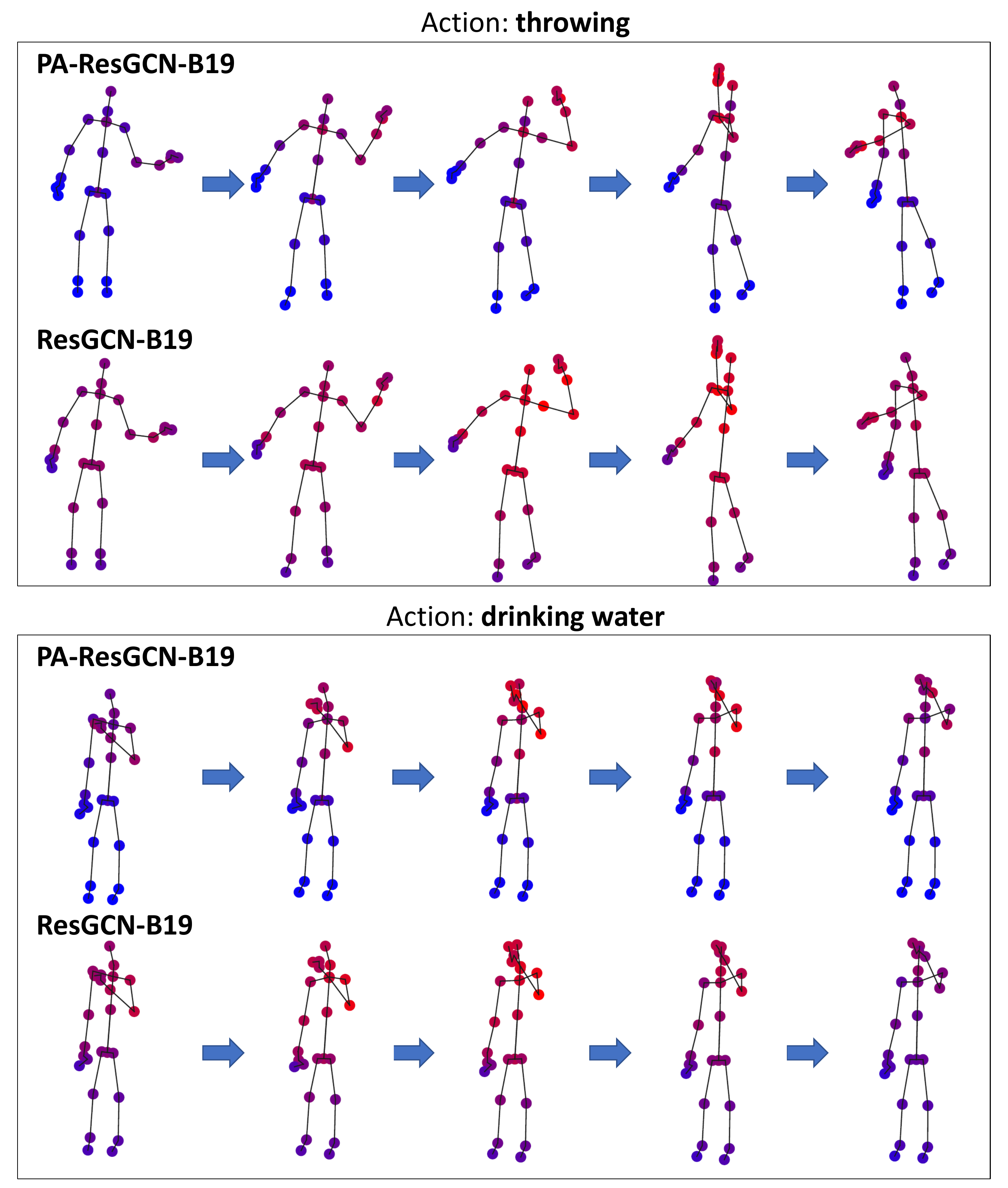}}
  \caption{Activated joints in several contextual frames of ResGCN-B19 and PA-ResGCN-B19 for the sample actions {\it throwing} and {\it drinking water}. The {\color{red}red} joints denote the activated joints, while {\color{blue}blue} means non-activated joints. \bv}\label{fig:activation}
  \Description{activation}
  \vspace{-0.4cm}
\end{figure}

\paragraph{Activation Map.} To show how our model works, the activation maps of some action sequences are calculated by class activation map technique \cite{learning2016zhou}, as presented in Fig. \ref{fig:activation}, in which the activated joints in several sampled frames are displayed. From this figure, we can find that the PA-ResGCN-B19 model successfully concentrates on the most informative body parts, \ie, left arm for the two actions. Besides, compared with the ResGCN-B19 model, the PA-ResGCN-B19 pays obviously higher attention to the left arm, while ResGCN-B19 shows nearly equal attention to the whole upper body. This significant difference implies that the proposed PartAtt block is more explainable than the joint-based methods.

\paragraph{Failure Cases.} Although PA-ResGCN receives promising results on the large-scale datasets, there are still two actions which are difficult to recognize (accuracies less than 70\%). They are the actions {\it reading} and {\it writing}. It is easy to find that, both the two actions are performed by two hands, and are extremely similar with each other. However, there are only two joints are recorded for each hand in the two datasets. Therefore, it is still challenging for our model to capture such subtle movements of two hands.

\section{Conclusion}
\label{sec:conclusion}

In this paper, we have proposed an efficient but strong baseline based on the MIB, residual bottleneck blocks and PartAtt blocks. Different from other attention enhanced models, the proposed PartAtt block concentrates more on the essential body parts, instead of joints, which makes the model avoid focusing on some superfluous even interferential features. In order to save the training and inference time, we utilize the bottleneck technique into the GCN model, which significantly reduces the number of learnable parameters, at most 34 times less than other models. On the challenging datasets, NTU RGB+D 60 \& 120, the proposed PA-ResGCN achieves the SOTA performance, while its inference speed is obviously higher than other models. Thus, the new baseline will have huge potential for some complex extensions. In the future, we will extend the proposed baseline with the object appearance, which is responsible for the recognition of the extremely similar actions.

\iffinal

\section*{Acknowledgement}
\label{sec:Acknowledgement}

This work is sponsored by National Key R\&D Program of China (No.2016YFB1001002), National Natural Science Foundation of China (No.61525306, No.61633021, No.61721004), Shandong Provincial Key Research and Development Program (Major Scientific and Technological Innovation Project) (No.2019JZZY010119) and CAS-AIR.

\fi

\bibliographystyle{ACM-Reference-Format}
\balance
\bibliography{conferences,acmart}

\end{document}